\pdfoutput=1

\documentclass[11pt]{article}

\usepackage{EMNLP2023}

\usepackage{times}
\usepackage{latexsym}
\usepackage{graphicx}
\usepackage{subfig}
\usepackage{booktabs}
\usepackage{multirow}

\usepackage{calc}
\newcounter{tablemajor}

\newcommand*\settablecounter[1]{\setcounter{tablemajor}{#1}}

\newcommand\blfootnote[1]{%
  \begingroup
  \renewcommand\thefootnote{}\footnote{#1}%
  \addtocounter{footnote}{-1}%
  \endgroup
}

\usepackage[T1]{fontenc}

\usepackage[utf8]{inputenc}

\usepackage{microtype}

\usepackage{inconsolata}

\title{Location-Aware Visual Question Generation with Lightweight Models}

\newcommand{\sun}[1]{{\color{blue}{\small\bf\sf [Sun: #1]}}}

\newcommand{\ie}{\textit{i}.\textit{e}.,\ }
\newcommand{\eg}{\textit{e}.\textit{g}.,\ }

\newcommand{\myfig}[1]{Figure \ref{#1}}
\newcommand{\mytable}[1]{Table \ref{#1}}

\newcommand{\mysecref}[1]{Section \ref{#1}}

\newcommand{\dotieconcat}[2]{
  \text{\raisebox{.8ex}{$\smallfrown$}}%
}

\makeatletter
\newcommand\dslfontsize{\@setfontsize\dslfontsize\@viipt\@viiipt}
\makeatother

\newcommand{\myparagraph}[1]{\noindent\textbf{#1.}}

\usepackage{color}
\usepackage{xcolor}

\author{
Nicholas Collin Suwono$^{1, 2}$\quad 
Chih Yao Chen$^1$\quad 
Tun Min Hung$^1$\quad \\
\textbf{Ting-Hao `Kenneth' Huang}$^3$\quad 
\textbf{I-Bin Liao}$^4$\quad 
\textbf{Yung-Hui Li}$^4$\quad 
\textbf{Lun-Wei Ku}$^1$\quad \\
\textbf{Shao-Hua Sun}$^2$\\
  Institute of Information Science, Academia Sinica$^1$ \quad
  National Taiwan University$^2$\\
  Pennsylvania State University$^3$ \quad
  Hon Hai Research Institute$^4$ \\
  \texttt{r10946021@ntu.edu.tw} \quad 
  \texttt{cyaochen@cs.unc.edu} \quad
  \texttt{hungtungming@gmail.com} \quad \\
  \texttt{txh710@psu.edu} \quad 
  \texttt{yunghui.li@foxconn.com} \quad  
  \texttt{ibin.liao@foxconn.com} \quad \\
  \texttt{lwku@iis.sinica.edu} \quad  
  \texttt{shaohuas@ntu.edu.tw}
  }
 
\begin{document}
\maketitle

\begin{abstract}

This work introduces a novel task, location-aware visual question generation (LocaVQG), which aims to generate engaging questions from data relevant to a particular geographical location. Specifically, we represent such location-aware information with surrounding images and a GPS coordinate. To tackle this task, we present a dataset generation pipeline that leverages GPT-4 to produce diverse and sophisticated questions. Then, we aim to learn a lightweight model that can address the LocaVQG task and fit on an edge device, such as a mobile phone. To this end, we propose a method which can reliably generate engaging questions from location-aware information. Our proposed method outperforms baselines regarding human evaluation (\eg engagement, grounding, coherence) and automatic evaluation metrics (\eg BERTScore, ROUGE-2). Moreover, we conduct extensive ablation studies to justify our proposed techniques for generating the dataset and solving the task.
\blfootnote{Project page can be found at \url{https://github.com/AcademiaSinicaNLPLab/LocaVQG}} 


\end{abstract}

\section{Introduction}


Driving is an integral part of our daily routines, playing a significant role in our lives. Whether commuting to work, running errands, or embarking on exciting adventures, we heavily rely on automobiles to get us from one place to another. 
Despite its undeniable convenience, driving requires constant focus on the road, the need to remain alert, and the mental strain of navigating through traffic.
Hence, staying behind the wheel after long working hours or during a long-distance trip can give rise to hazardous circumstances.
To combat this, passengers often engage in conversation to keep the driver awake and attentive.


Can we develop a system running on a lightweight device that automatically engages the driver in a conversation?
While initiating a conversation with general questions may not interest the driver, delving into driver-specific inquiries raises privacy concerns since it requires personal information.
Our key insight is to engage the driver in a conversation by posing questions based on the location-aware information, composed of  both the geographical coordinate of the car and surrounding visual perception represented by pictures captured by on-car cameras.
Such rich location-aware information allows for producing diverse and relevant questions, enabling a system to initiate an engaging conversation.

\begin{figure}[t]
    \centering
    \includegraphics[width=\linewidth]{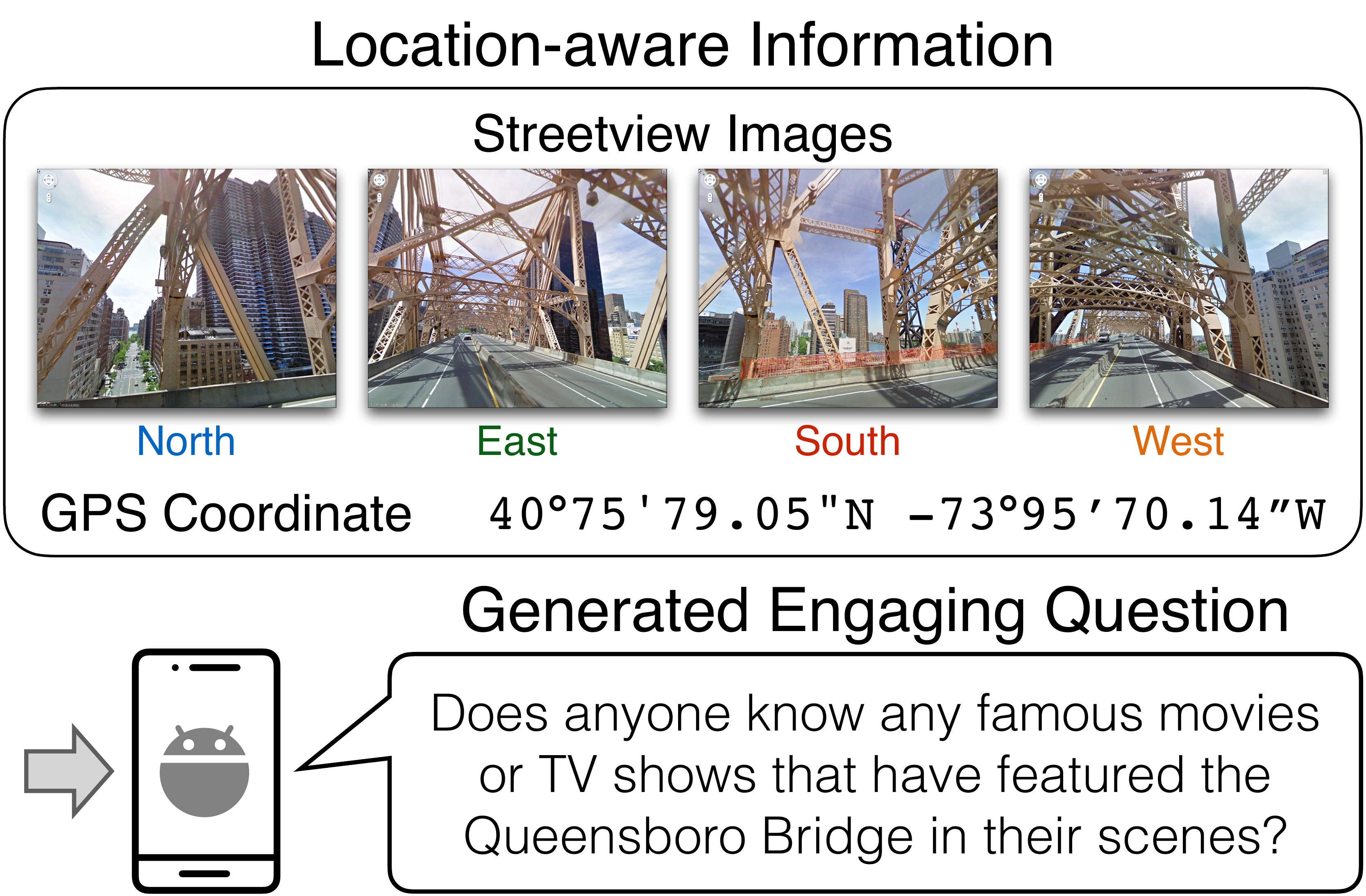}    
    \caption{
    \textbf{Location-aware Visual Question Generation (LocaVQG)} involves generating engaging questions from a specified location, represented by a GPS coordinate of a vehicle and a set of street view images captured by on-car cameras.
    }
    \label{fig:teaser}
\end{figure}

\begin{figure*}[t]
    \centering
    \includegraphics[width=\textwidth]{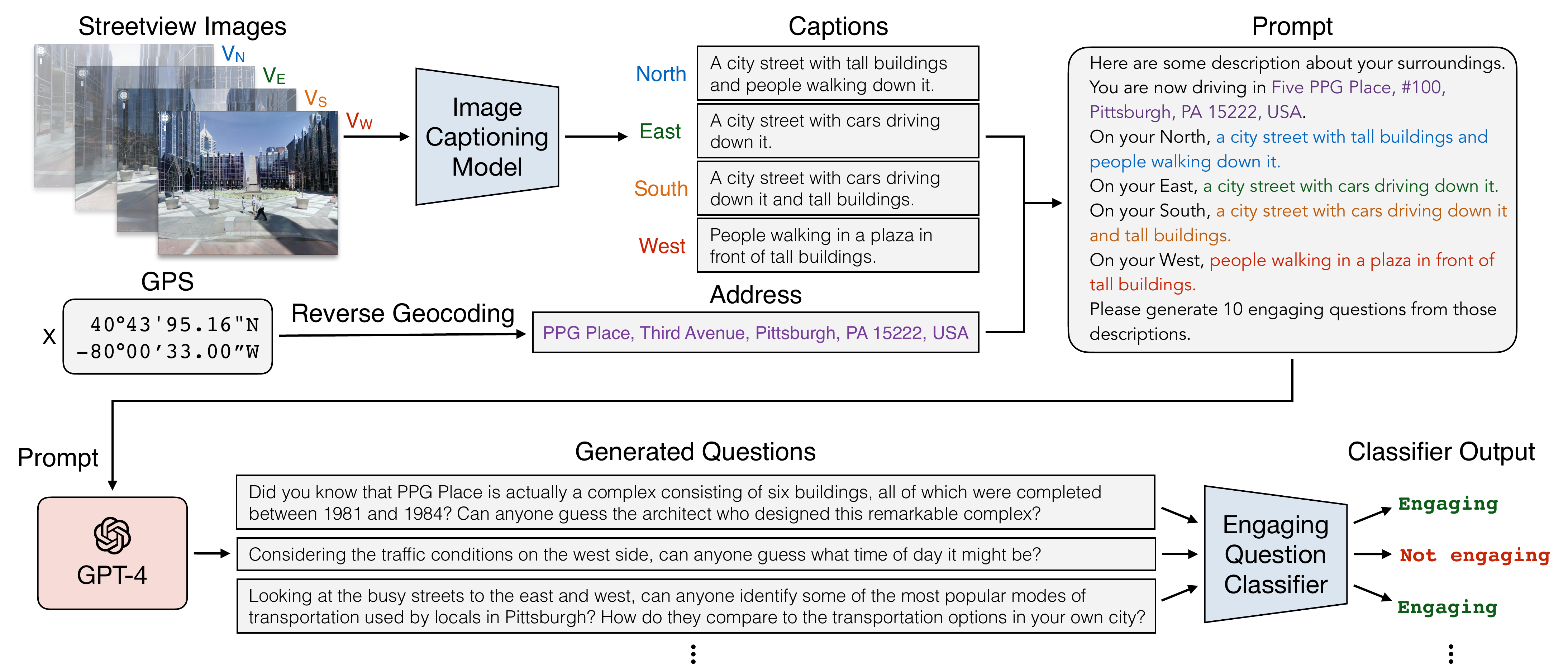}
    \caption{\textbf{Data Generation Pipeline}. 
    This pipeline produces questions from a given LocaVQG task tuple consisting of four street view images $V_N, V_E, V_S, V_W$ and a GPS coordinate $X$.
    We caption each image using an image captioning model and infer the address by reverse geocoding the GPS coordinate.
    Then, we construct a prompt that describes in detail the location-aware information and leverage GPT-4 to generate questions.
    We further employ an engaging question classifier to filter out non-engaging questions.
    Finally, the remained questions are included in the dataset with the given LocaVQG task tuple.
    }
    \label{fig:dataset_generation_pipeline}
\end{figure*}

In this work, we introduce a novel task, Location-aware Visual Question Generation (LocaVQG), which aims to produce engaging questions from a GPS coordinate of a vehicle and a set of street-view images captured by on-car cameras, as illustrated in~\myfig{fig:teaser}.
We make the first attempt to tackle this task by developing a data generation pipeline that can create a dataset containing high-quality samples for the LocaVQG task.
To this end, we leverage the recent advances in large language models (LLMs)~\cite{liu2023summary, touvron2023llama}. 
Specifically, we collect data from Google Street View and design a prompt according to the address obtained by reverse geocoding the GPS coordinate and the captions of street-view images provided by an off-the-shelf image captioning model.
While LLMs can generate questions relevant to the provided location-aware information, the produced questions may not always be engaging. 
Therefore, we further propose to train an engaging question classifier that can filter out non-engaging questions.
Our proposed dataset generation pipeline is illustrated in~\myfig{fig:dataset_generation_pipeline}.


We present a method, FDT5, that can learn a lightweight model and reliably address the LocaVQG task. 
We compare our proposed method to various small and mid-size language models learning from the generated dataset.
The experimental results demonstrate that our proposed FDT5 outperforms the baselines regarding human evaluation (\eg engagement, coherence, grounding) and automatic evaluation metrics, \eg BERTScore~\cite{DBLP:journals/corr/abs-1904-09675}, ROUGE-2~\cite{lin-2004-rouge}.
Our FDT5 with only $15$M parameters achieves competitive performance even compared to a large language model (\ie GPT-4).
This highlights the effectiveness of the proposed dataset generation pipeline as well as the proposed training techniques.


The main contributions of this work are three-fold as follows:
\begin{itemize}
    \item \myparagraph{Task} We propose Location-aware Visual Question Generation (LocaVQG), a novel task that aims to produce engaging questions from a GPS coordinate of a vehicle and a set of street-view images captured by on-car cameras. This will lead to the development of more intelligent in-car assistant systems.
    \item \myparagraph{Dataset} To address LocaVQG, we introduce a dataset generation pipeline that can 
    produce diverse and engaging questions from a specified location by leveraging pre-trained LLMs.
    \item \myparagraph{Method} 
    We present a method FDT5 that outperforms all the lightweight baselines regarding human evaluation (\eg engagement, coherence, grounding) and automatic evaluation metrics (\eg BERTScore, ROUGE-2).    
    
\end{itemize}

    \section{Related Works}


\label{sec-method}

\myparagraph{Self-driving cars}
Despite the recent advances in developing self-driving cars~\cite{electronics11142162}, most current commercialized autonomous vehicles are categorized as
SAE (the Society of Automotive Engineers)~\cite{sae2018taxonomy}
Level 2 (\eg Tesla, Hyundai, Kia) or Level 3 (\eg Mercedes).
When driving an SAE Level 2 vehicle, the driver must always hold the steering wheel.
With an SAE Level 3 vehicle, the driver must still be ready to take control of the vehicle at all times when prompted by the vehicle.
That being said, driving in modern days still requires the driver's attention and therefore can be assisted with the task and the system proposed in this work.

\myparagraph{In-car intelligent assistant system} 
Developing in-car intelligent assistant systems, 
such as a voice assistant~\cite{10.1145/3242587.3242593, 10.1145/3290605.3300270}, is an emerging research area.
\citet{driversfatigue} discovered that engaging drivers in a conversation could effectively reduce driver fatigue.
In contrast, this work focuses on raising drivers' attention by formulating a task and devising a system to produce location-aware engaging questions.

\myparagraph{Visual question generation (VQG)} 
VQG concerns generating questions from visual inputs~\cite{mostafazadeh-etal-2016-generating}. 
Compared to this work, recent works~\cite{lu-etal-2021-engage, MVQG} that explore VQG do not leverage geographical information (\eg GPS).
On the other hand, \citet{zhang2021situatedqa} presented a dataset with questions respecting geographical and temporal contexts; yet, it does not utilize visual inputs.
In contrast, the task and the dataset proposed in this work leverage images captured by on-car cameras.

\myparagraph{Large language models (LLMs)} 
Recent advances in LLMs~\cite{liu2023summary, touvron2023llama} have led to promising results in various domains~\cite{wei2022emergent, zhang2023bootstrap}.
However, these gigantic LLMs with billions of parameters~\cite{liu2023summary, touvron2023llama} cannot be deployed on lightweight devices and therefore are not ideal for in-car intelligent assistant systems. 
This work aims to develop lightweight models that can run on edge devices like mobile phones.

\myparagraph{Lightweight language models} 
Existing mobile-friendly language models \cite{sun2020mobilebert, mehta2022mobilevit} 
struggle at language generation tasks. 
This work aims to devise lightweight models that can address VQG tasks and achieve competitive results even compared to LLMs.

\section{LocaVQG: Location-aware Visual Question Generation}
\label{sec:task}

We introduce a novel task, Location-aware Visual Question Generation (LocaVQG).
This section formally defines this task and describes how we collect data to construct the LocaVQG task tuples.

\subsection{Location-aware Information} 

Location-aware Information includes data or content specifically relevant to, or influenced by a particular geographical location.
With such information, applications can offer location-specific recommendations, directions, local weather updates, nearby points of interest, targeted advertisements, etc.
Since our goal is to produce engaging questions with an in-car device based on location-aware information,
we limit it to the information that is easily accessible even without the internet.
Specifically, we consider the surrounding visual perception and the geographical coordinate.

\subsection{LocaVQG Task Tuple}

To collect the surrounding visual perception and the geographical coordinate of diverse locations, we propose to leverage Google Street View Dataset~\cite{6710175}.
The dataset contains 10,343 coordinates, and each coordinate comes with $5$ corresponding directional images (North, East, South, West, and Upper/Sky view). 
To ensure the location diversity, we select 3,759 coordinates with their $4$  directional images, excluding the upper/sky view, which is usually not observable by the driver.
We denote the geographical coordinate of each location as $X$ and its surrounding images as $V_d$ with $d=[N, E, S, W]$, standing for each direction.
We define our LocaVQG task tuple $T$ as: $T = [V_N, V_E, V_S, V_W, X]$.
Given a LocaVQG task tuple $T$, our goal is to produce an engaging question $Q$ with a model $f$: $f(T) = Q$.





\section{Generating LocaVQG Dataset}
\label{sec:dataset}

Our goal is to train lightweight models to address the
LocaVQG task.
Therefore, we aim to "label" the task tuples described in~\mysecref{sec:task}.
Annotating the task tuples with engaging questions requires creativity and location-specific domain knowledge, which can be challenging for human annotators.
In this work, we propose automatically generating questions from LocaVQG task tuples by leveraging the recent development of LLMs.
An overview of the proposed dataset generation pipeline is depicted in~\myfig{fig:dataset_generation_pipeline}.




\settablecounter{1}
\begin{table*}[t]
\centering
\scalebox{0.75}{\begin{tabular}{l}
    \toprule
    \textbf{Engaging Questions} \\
    \midrule
    \textcolor{teal}{The city of Pittsburgh is known for its numerous bridges}. How many bridges do you think are in the city, \\ and why do you think there are so many? \\
    What types of events or festivals might take place in \textcolor{blue}{this park} throughout the year? \\
    As we look towards the \textcolor{red}{south}, can you guess the purpose of this brick building with cars parked in front? Perhaps an office \\building, a restaurant, or something else? \\
    \midrule
    \midrule
    \textbf{Non-Engaging Questions} \\
    \midrule
    Speaking of the hospital, does anyone know the range of
    medical services provided at Prince George's Hospital? \\
    What are some ways that city planners might improve traffic
    flow at busy intersections? \\
    Noticing the mixture of architectural styles, can you guess which
    era had the most significant influence on the city's 
    development? \\
    \bottomrule
\end{tabular}}
\caption{\textbf{GPT-Generated Questions.} 
We provide examples of GPT-4 generated questions that are classified 
as engaging and non-engaging by the engaging question classifier. 
Answering these non-engaging questions often requires specific domain knowledge and therefore may interest only limited audience.
\textcolor{blue}{blue}-colored text indicate visual cues, \textcolor{red}{red}-colored text indicate directional cues, \textcolor{teal}{teal}-colored text indicate location-specific information.
}
\label{table:questions}
\end{table*}

\subsection{Prompting GPT-4} 
\label{sec:prompt}

This section describes how we utilize GPT-4~\cite{openai2023gpt4} to produce questions from LocaVQG task tuples by processing task tuples and designing LocaVQG prompts.

\myparagraph{Street view images $\rightarrow$ captions}
While GPT-4 is a multimodal model, its feature of taking image inputs is not yet publicly accessible as of May 2023.
Hence, to inform GPT-4 with the street view images,
we caption street view images using an off-the-shelf image captioning model~\cite{wang2022ofa}.

\myparagraph{GPS coordinate $\rightarrow$ address}
To leverage the GPS coordinate, we reverse geocode it using Google's Reverse Geocoding API~\cite{reversegeocode2023}, translating the coordinate into a street address. 
We found that with the decoded street address, GPT-4 can often infer nearby famous landmarks, or information, and generate relevant questions.



\myparagraph{Constructing prompts}
We aim to prompt GPT-4 with the processed location-aware input and produce engaging questions.
We first design a \textbf{system prompt} that infuses GPT-4 with a tour guide role, enforcing it to engage users.
Then, we design a \textbf{chat prompt} that encapsulates processed location-aware information and requires GPT-4 to generate engaging questions.
The two prompts are presented as follows.
\begin{itemize}
    \item \textbf{System prompt}: \textit{You are a tour guide and you are driving in a car with your tourists. You want to engage with them with any kind of information you have around you.}
    \item \textbf{Chat prompt}: \textit{Here are some descriptions of your surroundings You are currently driving on [Street Address]. On your North, [Image Caption]. On your East, [Image Caption]. On your South, [Image Caption]. On your West, [Image Caption]. Based on those descriptions, please ask 10 engaging questions.}
\end{itemize}

\subsection{Filtering GPT-Generated Questions} 
\label{sec:filter}

While GPT-4 can generate numerous diverse questions from our designed prompts, we empirically find that some generated questions are not particularly engaging (\eg requiring domain knowledge), as shown in~\mytable{table:questions}.
To combat this, we propose to learn a BERT-based~\cite{devlin2019bert} engaging question classifier to filter out non-engaging questions.
We construct the training data for this classifier with non-engaging questions from SQuaD~\cite{rajpurkar2016squad} 
and engaging questions from MVQG~\cite{MVQG}.
The key insight is that SQuaD questions are made for question-answering tasks, thus, solely revolves around a passage, while MVQG questions are collected with engagement in mind.

With this trained engaging question classifier, for each LocaVQG task tuple $T$, we filter out non-engaging questions generated by GPT-4, and the remained questions are included in the dataset as the "labels" for this task tuple.


\settablecounter{2}
\subsection{Dataset Statistics}
\label{sec:stats}
Applying the procedures described in~\mysecref{sec:prompt} and~\mysecref{sec:filter} 
results in a dataset with $3759$ task tuples and $35$K questions.
The basic statistics of the dataset are described in~\mytable{table:stats}.

\begin{table}[h]
\centering
%
\scalebox{0.75}{\begin{tabular}{l l}
    \toprule
    \# of LocaVQG Task Tuples & 3759\\
    - \# of Task Tuples from Pittsburgh & 919\\
    - \# of Task Tuples from Orlando & 611\\
    - \# of Task Tuples from New York & 2217\\
    \# of Questions After Filtering & 35551 \\
    Average Sentence Length &  16.6 \\
    Average Question Length &  30.8 \\
    \bottomrule
\end{tabular}
}
\caption{\textbf{Dataset Statistics.} 
We present the statistics of our location-aware visual question generation dataset
}
\label{table:stats}
\end{table}

\subsubsection{Question Length}

\begin{figure}[t]
    \centering
    \includegraphics[width=0.64\linewidth]{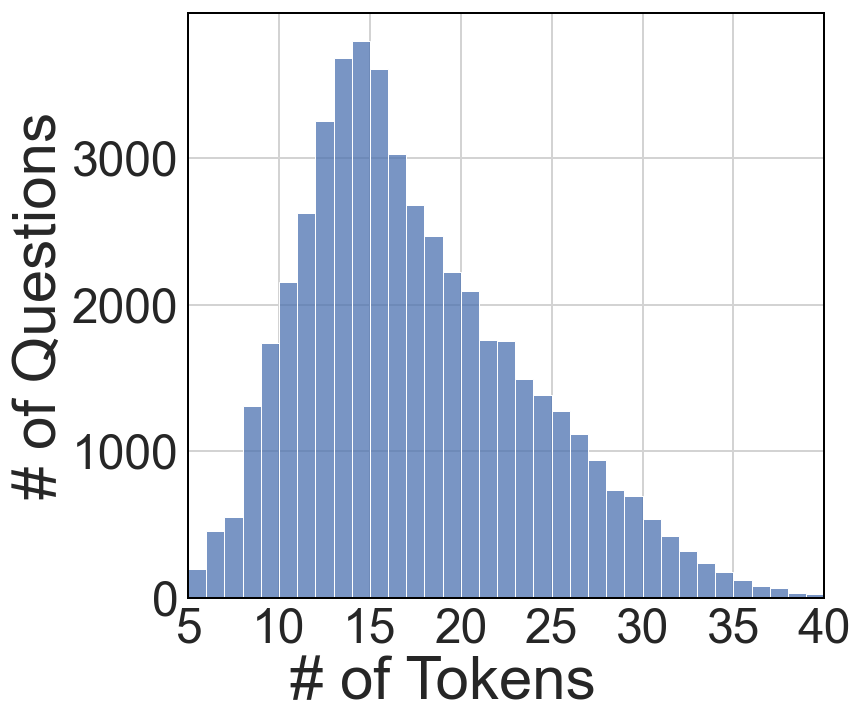}
    \includegraphics[width=0.32\linewidth]{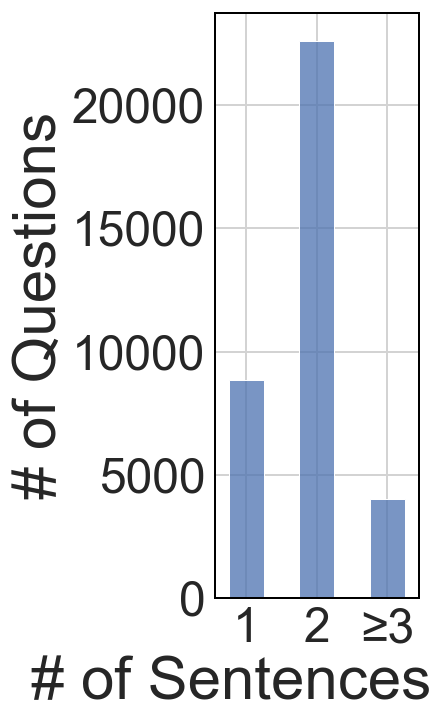}
    \caption{\textbf{Question Length.}
    We present the question length statistics in \# and tokens and \# of sentences.
    }
    \label{fig:hist_q_len}   
\end{figure}

We present the histograms of question lengths in terms of \# of tokens and \# of sentences in~\myfig{fig:hist_q_len}.

\subsubsection{Frequent Trigrams and Words}

We present the top $15$ frequent trigrams in the dataset in~\mytable{table:frequent_trigram}.
The questions often start by trying to intrigue the respondent (\eg Did you know, What do you). 
Also, open-ended questions (\eg \textit{have you noticed}, \textit{have you ever}) appear quite frequently. 
Most frequent words, presented in~\mytable{table:frequent_words},
require or lead the attention of the respondent (\eg \textit{considering}, \textit{looking}, \textit{notice}). 

\settablecounter{3}

\begin{table}[h]
\centering
\scalebox{0.75}{
\begin{tabular}{l l l}
    \toprule
    Did you know & What do you & Can you spot   \\
    Can you guess & As we drive & How do you    \\
    Have any of & Have you noticed & Can anyone guess \\
    Have you ever & Can you identify & What are your  \\
    Are you familiar & What are some & As we continue  \\
    \bottomrule
\end{tabular}}
\caption{\textbf{Top 15 Frequent Trigram of Questions.} 
The frequent trigrams appearing in the dataset show that the questions aim to intrigue or engage the respondent.}
\label{table:frequent_trigram}
\end{table}

\settablecounter{4}

\begin{table}[h]
\centering
\scalebox{0.75}{
\begin{tabular}{l l l l l}
    \toprule
    Speaking & Considering & Based & Looking & Notice  \\
    Since & Perhaps & Observing & Residential & New   \\
    Look & Pittsburgh & Given & Orlando & See  \\
    Let & Turning & Judging & Noticing & Besides \\
    \bottomrule
\end{tabular}}
\caption{\textbf{Top 20 Frequent Words.} This table presents 20 frequent words appearing in the dataset aside from commonly used words, such as "\textit{have}", "\textit{can}".}
\label{table:frequent_words}
\end{table}


\subsubsection{Question Quality}

We compare our proposed dataset to an engaging question dataset, the MVQG dataset~\cite{MVQG}, regarding the following criteria and report the results in~\mytable{table:comparison_mvqg}.

\begin{itemize}
    \item \textbf{Vocabulary Size (Vocab Size)} measures the number of distinct words in a dataset.
    
    \item \textbf{Average Sentence Length~(Avg Sent. Length)} computes the average length of sentences across the whole dataset, representing how rich and detailed a dataset is.
    
    \item \textbf{Syntactic Complexity} calculates 
    the degree of variation, sophistication, and elaboration of the questions in a dataset~\cite{ferraro-etal-2015-survey}. 
    We report the mean of \textbf{Yngve Score} normalized by the sentence length. 
    
    \item \textbf{Percentage of Abstract Terms (\% Abstract Term)} computes the ratio of visual and non-visual concepts covered by a dataset,
    based on the abstract terms defined by \citet{vanderwende-etal-2015-amr}.
    
    \item  \textbf{Average Term Depth} is calculated based on WordNet, where noun words with a lower depth indicate higher-level concepts \cite{lu-etal-2021-engage}.
\end{itemize}

Compared to the MVQG dataset, the results show that our proposed LocaVQG dataset contains significantly more diverse and sophisticated questions.
In fact, the questions included in the MVQG dataset are collected from human annotators.
This highlights the effectiveness of generating questions by leveraging the recent advances in LLMs (\eg GPT-4), as proposed in this work. 
Further evaluations on the generated questions can be found in~\mysecref{sec:app_address} and~\mysecref{sec:app_visual_learning}.


\settablecounter{5}

\begin{table}[h]
\centering
\scalebox{0.75}{%
\begin{tabular}{l c c}
    \toprule
    \textbf{Criteria} & \textbf{MVQG} & \textbf{LocaVQG (Ours)} \\
    \midrule
    Vocabulary Size $\uparrow$ & 608 & \textbf{3046}  \\ 
    Average Sentence Length $\uparrow$ & 12.341 & \textbf{17.168} \\ 
    Yngve Score $\uparrow$ & 2.271 & \textbf{3.761} \\
    \% Abstract Terms $\uparrow$ & 0.127 & \textbf{0.167} \\
    Average Term Depth $\downarrow$ & 7.906 & \textbf{7.259} \\
    \bottomrule
\end{tabular}}
\caption{\textbf{Question Quality Comparison with MVQG}. 
}
\label{table:comparison_mvqg}
\end{table}

\section{Learning and Evaluating Lightweight Models}
\label{sec:model}

We aim to train and evaluate lightweight models learning from the proposed LocaVQG dataset.

\settablecounter{6}

\begin{table*}[t]
\centering
\scalebox{0.75}{\begin{tabular}{l c c c c c c | c}
    \toprule
    \textbf{Model} & \textbf{\#Parameters} & \textbf{Engagement} & \textbf{Naturalness} & \textbf{Coherence} & \textbf{Common Sense} & \textbf{Grounding} & \textbf{Overall} \\
    \midrule
    MVQG-VL-T5 & 254M & 3.84 & 3.64 & 3.65 & 3.81 & 3.84 & 3.76   \\
    \midrule
    MVQG-VL-T5$_{fine-tuned}$ & 254M & 3.96 & 3.82 & 3.82 & 3.99 & 3.66 & 3.85   \\
    T5-Large & 770M  & 3.92 & 3.81 & 3.78 & 4.03 & 3.83 & 3.87\\ 
    T5-Base & 220M & 3.92 & 3.81 & 3.73 & 3.97 & 3.78 & 3.84\\ 
    T5-Tiny & \textbf{15.6M} & 3.96 & 3.79 & 3.67 & 4.01 & 3.81 & 3.85 \\ 
    FDT5 (Ours) & \textbf{15.6M} & \textbf{4.03} & \textbf{3.83} & \textbf{3.96} & \textbf{4.05} & \textbf{4.03} & \textbf{3.98}\\
    \midrule
    GPT-4 & 1T$^\ast$ & 4.12 & 3.99 & 4.01 & 4.05 & 4.01 & 4.04\\
    \midrule  
    Human Annotator & - &  4.06 & 3.87 & 3.90 & 4.06 & 3.88 & 3.95\\
    \bottomrule
\end{tabular}}
\caption{\textbf{Human Evaluations.} Each question is rated by three AMT workers. Among all the light-weight models, our proposed FDT5 achieves the best overall performance and has the fewest parameters.
Note that while the exact number of parameters GPT-4 is not revealed, many believe it is at least 6 times larger than GPT-3~\cite{brown2020language} ($175$B).
}
\label{table:human_eval}
\end{table*}

\subsection{Baselines}

We compare our method to the following baselines.

\myparagraph{Text-To-Text Transfer Transformer (T5)} 
We experiment with a family of T5 pre-trained language models~\cite{raffel2020exploring, tay2022scale}, which includes T5-Large ($770$M), T5-Base ($220$M), and T5-Tiny ($15.6$M).
We fine-tune the pre-trained T5 models on our LocaVQG dataset.
Specifically, for each LocaVQG task, the models learn to map the prompt presented in~\mysecref{sec:prompt} to one of the ground truth questions generated by GPT-4.

\myparagraph{MVQG-VL-T5} 
\citet{cho2021vlt5} introduced Vision-and-Language T5 (VLT5) for vision-and-language tasks. 
\citet{MVQG} adapted it for generating questions from a set of images.
We adopt this method, dubbed MVQG-VL-T5, and fine-tune the pre-trained model on our LocaVQG dataset.
The input of MVQG-VL-T5 consists of $4$ street view images and the street address.

More details can be found in~\mysecref{sec:app_model}.

\subsection{Our Approach}
\label{sec:dt5t}

We propose Filtered Distilled T5-Tiny (FDT5).

\myparagraph{Distillation} While T5-Tiny has the fewest of parameters and can fit on mobile phones, 
its capacity might be limited to a complex task like LocaVQG.
Therefore, we propose to learn a T5-Tiny model by distilling a learned T5-Large model.
Inspired by~\citet{DBLP:journals/corr/abs-1911-03829},
during training, we utilize both the questions generated by GPT-4 from the dataset and the questions generated by the T5-Large model, resulting in the objective:
\begin{equation}
    \mathcal{L}(\theta)= \alpha \cdot \mathcal{L}_{hard}(\theta) + (1 - \alpha) \cdot \mathcal{L}_{soft}(\theta),
\end{equation}
where $\alpha$ balances the relative importance of learning from each loss and $\theta$ parameterizes the model. The hard-label loss $\mathcal{L}_{hard}$ (ground truth target) optimizes cross-entropy, while the soft-label loss $\mathcal{L}_{soft}$ (teacher model) optimizes KL Divergence.

\myparagraph{Filtering} 
To further improve the engagingness of the questions produced by our method,
we propose to utilize the engaging questions classifier described in~\mysecref{sec:filter}
to filter out non-engaging questions.
Specifically, given a LocaVQG task, 
our method keeps generating questions until accepted (\ie classified as "engaging") by the classifier.

Our proposed method FDT5 combines the two techniques described above.

\subsection{Human Evaluation}
\label{sec:human_eval}
We provide human evaluation of the questions generated by all the methods.

\subsubsection{Evaluation Metrics}

We randomly sampled $100$ LocaVQG task tuples and the questions produced by all the models. 
Each question is evaluated by three Amazon Mechanical Turk (AMT) workers according to the following metrics.
We adopt a 5-point Likert scale for all the evaluation metrics.
\begin{itemize}
    \item \textbf{Engagement}: You find the question engaging and you would want to answer the question.
    \item \textbf{Naturalness}: It is natural to ask this question given the information you have.
    \item \textbf{Coherence}: The question is coherent with the information you have.
    \item \textbf{Common Sense}: It makes sense to ask these questions given the information you have.
    \item \textbf{Grounding}: The question asked about things related to the information you have.
\end{itemize}

\settablecounter{8}

\begin{table*}[t]
\centering
\scalebox{0.75}{\begin{tabular}{l c c c c c | c}
    \toprule
    \textbf{Model} & \textbf{Engagement} & \textbf{Naturalness} & \textbf{Coherence} & \textbf{Common Sense} & \textbf{Grounding} & \textbf{Overall} \\
    \midrule
    Filtered Dataset (Ours) & \textbf{3.92} & \textbf{3.81} & 3.73 & \textbf{3.97} & \textbf{3.78} & \textbf{3.84}\\
    Unfiltered Dataset & 3.89 & 3.76 & \textbf{3.78} & 3.85 & \textbf{3.78} & 3.81 \\ 
    \bottomrule
\end{tabular}}
\caption{\textbf{Engaging Question Classifier for Dataset Generation}. 
Employing the engaging question classifier in the dataset generation process to filter out unengaging questions improves the quality of generated questions.
}
\label{table:ablation_filter}
\end{table*}

\settablecounter{9}

\begin{table*}[t]
\centering
\scalebox{0.75}{\begin{tabular}{l c c c c c | c}
    \toprule
    \textbf{Model} & \textbf{Engagement} & \textbf{Naturalness} & \textbf{Coherence} & \textbf{Common Sense} & \textbf{Grounding} & \textbf{Overall} \\
    \midrule
    Filtered Inference (Ours) & \textbf{4.03} & \textbf{3.83} & \textbf{3.96} & \textbf{4.05} & \textbf{4.03} & \textbf{3.98}\\
    Unfiltered Inference & 3.96 & 3.78 & 3.82 & 4.04 & 3.79 & 3.88 \\ 
    \bottomrule
\end{tabular}}
\caption{\textbf{Engaging Question Classifier for Inference}. 
Our proposed FDT5 employs the classifier during inference to filter out unengaging questions. Excluding the filtering phase results in significantly worse performance.
}
\label{table:ablation_post_filter}
\end{table*}

We also provide the evaluation of the questions generated by GPT-4, which can be considered as an upper bound as GPT-4 has an unparalleled number of parameters compared to the lightweight models.
Furthermore, to compare the performance of these LMs against humans, we crowdsource and collect $100$ questions on AMT based on the same set of LocaVQG task tuples.
More details on AMT can be found in~\mysecref{sec:app_amt}.



\subsubsection{Results}

The human evaluations are presented in~\mytable{table:human_eval}.

\myparagraph{FDT5 outperforms all the lightweight models}
Our proposed method FDT5 achieves the best overall score with the fewest parameters.
This justifies the effectiveness of our adopted distillation scheme.
Furthermore, an average score of $3.98$ indicates that our model can reliably generate satisfactory questions from location-aware information.

\myparagraph{MVQG-VL-T5}
MVQG-VL-T5, without learning from our dataset, achieves the worst performance, demonstrating the importance of constructing and learning from a dataset dedicated to the LocaVQG task. Alternatively, the MVQG-VL-T5 model fine-tuned on our dataset (MVQG-VL-T5$_{fine-tuned}$) struggles at grounding, aligning with the findings discussed in~\cite{MVQG}.



\myparagraph{GPT-4 asks better questions than humans} 
The questions produced by GPT-4 are preferred by the workers compared to those provided by human annotators on all the metrics, except for common sense.
This justifies our proposed dataset generation pipeline, which collects questions from GPT-4 instead of humans.


\subsection{Automatic Evaluation Metrics}

\settablecounter{7}

\begin{table}[h]
\centering
\resizebox{\linewidth}{!}{%
\begin{tabular}{l c c c c c}
    \toprule
    \textbf{Model} & \textbf{BLEU-4} & \textbf{ROUGE-2} & \textbf{BERTScore} & \textbf{BLEURT} \\
    \midrule
    VLT5 & 0.2712  & 0.0342 & 0.5093 & \underline{-0.7208}    \\
    T5-Large & \textbf{0.2756}  & 0.0380 & \underline{0.5165} & -0.7336   \\ 
    T5-Base & \underline{0.2746}  & \underline{0.0388} & 0.5163 & -0.7305   \\ 
    T5-Tiny & 0.2635  & 0.0371 & 0.5164 & -0.7419   \\ 
    FDT5 (Ours) & 0.2661  & \textbf{0.0393} & \textbf{0.5190} & \textbf{-0.7073}  \\
    \bottomrule
\end{tabular}}
\caption{\textbf{Automatic Evaluation.} Our proposed FDT5 achieves the best performance on 3 out of 4 metrics (\ie ROUGE-2, BERTScore, and BLEURT).
}
\label{table:auto_eval}
\end{table}

We further evaluate the questions generated by all the models with some automatic evaluation metrics. 
To compare two questions based on exact wording, we are using BLEU-4~\cite{papineni-etal-2002-bleu}, ROUGE-2~\cite{lin-2004-rouge}. 
To compare the questions based on semantic similarity, we are using ML-Based evaluation: BERTScore~\cite{DBLP:journals/corr/abs-1904-09675}, and BLEURT~\cite{DBLP:journals/corr/abs-2004-04696}.
The results are presented in~\mytable{table:auto_eval}.
Our proposed method FDT5 achieves the best performance on ROUGE-2, BERTScore, and BLEURT.
T5-Large outperforms others in terms of BLEU-4.
This verifies the effectiveness of the filtering and distillation technique employed in FDT5.



\subsection{Ablation Study}

We conduct extensive ablation studies to investigate 
the effectiveness of employing the filtering classifier (\mysecref{sec:ablation_filter}), 
the impact of incorporating GPS coordinates into the question generation process (\mysecref{sec:ablation_address}),
and the effect of varying dataset sizes (\mysecref{sec:ablation_dataset_size}).


\begin{figure*}[t]
    \centering
    \includegraphics[width=\textwidth]{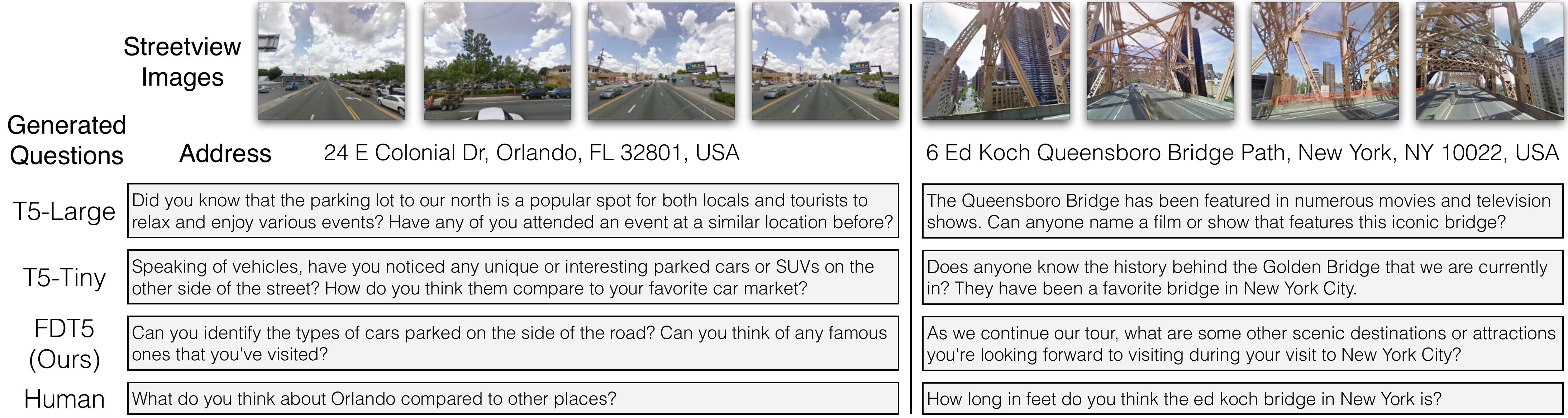}
    \caption{\textbf{Qualitative Results}. 
    We present sampled generated questions from T5-Large, T5-Tiny, our proposed method FDT5, and human annotators, 
    together with corresponding streetview images and addresses.
    With only $15$M parameters, FDT5 can reliably generate engaging location-aware questions.
    }
    \label{fig:qual}
\end{figure*}

\subsubsection{Employing Engaging Question Classifier}
\label{sec:ablation_filter}

We propose to learn an engaging question classifier for 
(1) filtering out non-engaging questions generated by GPT-4 during the dataset generation process (\mysecref{sec:filter}), and
(2) filtering out non-engaging questions produced by our model during inference (\mysecref{sec:dt5t}).
This section examines the effect of employing this classifier.

\myparagraph{Dataset Generation}
To verify the effectiveness of filtering out questions from GPT-4 with the classifier, we train a T5-Base model to learn from an unfiltered dataset that contains all the questions produced by GPT-4 (Unfiltered Dataset).
We compare the performance of this model to the T5-Base model trained on our proposed filtered dataset (Filtered Dataset) and report the human evaluation results in~\mytable{table:ablation_filter}.
The results demonstrate that the model learning from the filtered dataset achieves better performance, justifying the efficacy of employing the classifier.


\myparagraph{Inference}
We propose to filter out non-engaging questions generated during inference, adopted in our method FDT5.
We conduct human evaluations on filtered generation questions (Filtered Inference) and unfiltered questions (Unfiltered Inference), reported in~\mytable{table:ablation_post_filter}.
The results show that filtering non-engaging generated questions with the classifier can significantly improve the question quality on all the metrics.
This justifies the effectiveness of employing the classifier during inference.




\subsubsection{Incorporating GPS Coordinates}
\label{sec:ablation_address}

While~\citet{MVQG} explored generating questions from a set of images, our work further incorporates addresses (reverse geocoded from GPS coordinates) into the question generation process.
This section investigates the effect of employing such information.
We compare the questions generated by GPT-4 with or without the address in the prompt and report the results in~\mytable{table:ablation_address}.
The results show that incorporating the address leads to richer and more diverse questions, verifying the unique value of the proposed LocaVQG task.


\settablecounter{10}

\begin{table}[h]
\centering
\resizebox{\linewidth}{!}{%
\begin{tabular}{l c c}
    \toprule
    \textbf{Criteria} & \textbf{w/o address} & \textbf{w/ address (ours)} \\
    \midrule
    Vocabulary Size $\uparrow$ & 450 & \textbf{525}  \\ 
    Average Question Length $\uparrow$ & 25.02 & \textbf{30.18} \\ 
    Yngve Score $\uparrow$ & 3.531 & \textbf{3.693} \\
    \bottomrule
\end{tabular}}
\caption{\textbf{Effect of Leveraging Street Address} The questions generated with street addresses are richer and more diverse. }
\label{table:ablation_address}
\end{table}


\subsubsection{Varying Dataset Sizes}
\label{sec:ablation_dataset_size}

We investigate the impact of varying dataset sizes with T5-Tiny and our proposed FDT5, and report the results n~\mytable{table:ablation_dataset_size}.
FDT5 achieves better performance with fewer data points and performs comparably to T5-Tiny when dataset size increases.
This indicates that our method is more data efficient.


\settablecounter{11}

\begin{table}[h]
\centering
\scalebox{0.63}{\begin{tabular}{c c c c c c c}
        \toprule
        \textbf{Model} & \textbf{\#Samples} & \textbf{BLEU-4} & \textbf{ROUGE-2} & \textbf{BERTScore} & \textbf{BLEURT} \\        
        \midrule
        \multirow{5}{*}{T5-Tiny}
        & 0.7K & 0.2566 & 0.0366 & 0.5160 & -0.7666   \\
        & 1.7K & 0.2629  & 0.0341 & 0.5139 & -0.7530   \\ 
        & 2.7k & 0.2604  & \textbf{0.0374} & \underline{0.5164} & -0.7589   \\ 
        & 3.7K & \underline{0.2635}  & \underline{0.0371} & \textbf{0.5156} & \underline{-0.7419}  \\ 
        & 4.7K & \textbf{0.2639}  & 0.0361 & 0.5145 & \textbf{-0.7398}  \\ 
        \midrule
        \multirow{5}{*}{\shortstack{FDT5 \\(Ours)}}
        & 0.7k & 0.2565  & 0.0361 & 0.5201 & -0.7149  \\ 
        & 1.7k & \underline{0.2700}  & \underline{0.0402} & \textbf{0.5214} & -0.7245  \\ 
        & 2.7k & 0.2675  & \textbf{0.0422} & \underline{0.5211} & \underline{-0.7126}  \\ 
        & 3.7k & 0.2661  & 0.0393 & 0.5190 & \textbf{-0.7073}  \\
        & 4.7k & \textbf{0.2706}  & 0.0386 & 0.5180 & -0.7256 \\
        \bottomrule
\end{tabular}}
\caption{\textbf{Effect of Dataset Size.} From the results, FDT5 is more data efficient as it could achieve better performance with smaller sample size}
\label{table:ablation_dataset_size}
\end{table}

\subsection{Qualitative Results}

As human evaluations can be subjective, we present qualitative results in~\myfig{fig:qual} for the readers to better understand the generated questions.
The results show that FDT5 with only $15$M parameters can reliably generate engaging location-aware questions.





\section{Conclusion}
In this work, we propose a novel task, location-aware visual question generation (LocaVQG),
which aims to generate engaging questions from data relevant to a particular geographical location. 
Specifically, we represent location-aware information using four directional street view images and a GPS coordinate. 
To address this task,
we introduce a dataset generation pipeline that 
leverages the recent advances of large language models (\ie GPT-4) to generate diverse and sophisticated questions.
To ensure the engagingness of the questions produced by GPT-4,
we employ an engaging question classifier to filter out non-engaging questions.
Our proposed dataset contains richer and various questions compared to existing datasets.

To learn from the proposed LocaVQG dataset with lightweight models,
we present Filtered Distilled T5-Tiny (FDT5) method.
We extensively evaluate the proposed method and the baselines 
with human evaluation and automatic evaluation metrics.
Our proposed FDT5, with the fewest parameters, demonstrates superior performance on most metrics.
We conduct extensive ablation studies to verify the effect of employing the filtering classifier, the effectiveness of incorporating GPS coordinates into the question generation process, and the impact of varying dataset sizes.
We hope this work will encourage researchers to explore the LocaVQG task and its applications.



\section*{Limitations}

We discuss the limitations and how we can potentially address them in this section.

\myparagraph{Biases in AMT workers}
We notice that the AMT workers involved in human evaluation might be biased due to their demographic. This can potentially be addressed by ensuring the diversity of their background.

\myparagraph{Location-aware information}
Aiming to develop an in-car intelligent assistant, this work proposes representing location-aware information as a GPS coordinate and a set of images captured by on-car cameras.
Incorporating more detailed information, such as local news and weather, can potentially lead to more diverse and engaging questions,
and is left for future research.

\myparagraph{Address-aware LLMs}
Our proposed dataset generation pipeline heavily relies on
GPT-4.
This partially limits the generated questions to locations/addresses that are known by GPT-4 and therefore this pipeline might not produce coherent questions given locations that are less known by GPT-4.
We can potentially address this by employing a more sophisticated external information retrieval system to extract information from a location

\myparagraph{Human evaluation setup}
While our motivation is to develop an in-car intelligent system that can engage a driver in a conversation to keep the driver awake, this work falls short from the following perspectives. First, our work solely focuses on generating a question without considering continuing a conversation. Second, we evaluate the generated questions with an AMT interface where the AMT workers read and evaluate the questions. However, in a driving scenario, interacting with a virtual assistant by reading a question is impractical. Hence, evaluating the generated questions by connecting to a text-to-speech system and requiring the annotators to rate the questions by listening to them would align better with the application.

 \myparagraph{Distractingness of generated questions}
This work makes the very first attempt to develop an in-car visual question generation system that can ask engaging questions to initiate conversations with drivers.
However, such engaging questions can potentially distract drivers and lead to dangerous situations in the worst case.
To address such a concern,
we encourage future works along this line to consider the “distractingness” of generated questions. 
In particular, developing evaluation metrics to determine if a question would distract a driver from the road and devising methods to produce engaging yet non-distracting questions are promising and interesting research directions.




\section*{Ethics Statement}

Since our proposed dataset generation pipeline involves collecting questions from GPT-4, the data inherits any biases of GPT-4.
Moreover, our proposed method learns from the dataset,
and therefore will also be biased.
Therefore, inheriting biases can lead to
generating inappropriate, sexist, racist questions or descriptions. 
Fortunately, addressing ethical issues has been an active research area~\cite{liang2021towards, baldini2021your, yan2023practical, kasneci2023chatgpt}.
We wish to incorporate the advances in the field to alleviate ethical concerns.

\section*{Acknowledgement}

This work is supported by the National Science and Technology Council of Taiwan under grants 111-2634-F-002-022- and by the Academia Sinica and Hong Hai Research Institute collaborative project 05T-1110523-1Q.
Shao-Hua Sun was partially supported by the National Taiwan University and its Department of Electrical Engineering, Graduate Institute of Communication Engineering, and College of Electrical Engineering and Computer Science, and the Yushan Fellow Program by the Ministry of Education, Taiwan. 
We thank the online crowd workers for participating in the study.




\bibliography{emnlp2023}

\begin{thebibliography}{41}
\expandafter\ifx\csname natexlab\endcsname\relax\def\natexlab#1{#1}\fi

\bibitem[{Baldini et~al.(2021)Baldini, Wei, Ramamurthy, Yurochkin, and
  Singh}]{baldini2021your}
Ioana Baldini, Dennis Wei, Karthikeyan~Natesan Ramamurthy, Mikhail Yurochkin,
  and Moninder Singh. 2021.
\newblock Your fairness may vary: Pretrained language model fairness in toxic
  text classification.
\newblock \emph{arXiv preprint arXiv:2108.01250}.

\bibitem[{Braun et~al.(2019)Braun, Mainz, Chadowitz, Pfleging, and
  Alt}]{10.1145/3290605.3300270}
Michael Braun, Anja Mainz, Ronee Chadowitz, Bastian Pfleging, and Florian Alt.
  2019.
\newblock At your service: Designing voice assistant personalities to improve
  automotive user interfaces.
\newblock In \emph{CHI Conference on Human Factors in Computing Systems}.

\bibitem[{Brown et~al.(2020)Brown, Mann, Ryder, Subbiah, Kaplan, Dhariwal,
  Neelakantan, Shyam, Sastry, Askell, Agarwal, Herbert-Voss, Krueger, Henighan,
  Child, Ramesh, Ziegler, Wu, Winter, Hesse, Chen, Sigler, Litwin, Gray, Chess,
  Clark, Berner, McCandlish, Radford, Sutskever, and
  Amodei}]{brown2020language}
Tom~B. Brown, Benjamin Mann, Nick Ryder, Melanie Subbiah, Jared Kaplan,
  Prafulla Dhariwal, Arvind Neelakantan, Pranav Shyam, Girish Sastry, Amanda
  Askell, Sandhini Agarwal, Ariel Herbert-Voss, Gretchen Krueger, Tom Henighan,
  Rewon Child, Aditya Ramesh, Daniel~M. Ziegler, Jeffrey Wu, Clemens Winter,
  Christopher Hesse, Mark Chen, Eric Sigler, Mateusz Litwin, Scott Gray,
  Benjamin Chess, Jack Clark, Christopher Berner, Sam McCandlish, Alec Radford,
  Ilya Sutskever, and Dario Amodei. 2020.
\newblock Language models are few-shot learners.
\newblock \emph{arXiv preprint arXiv:2005.14165}.

\bibitem[{Chen et~al.(2020)Chen, Gan, Cheng, Liu, and
  Liu}]{DBLP:journals/corr/abs-1911-03829}
Yen{-}Chun Chen, Zhe Gan, Yu~Cheng, Jingzhou Liu, and Jingjing Liu. 2020.
\newblock Distilling the knowledge of {BERT} for text generation.
\newblock In \emph{ACL}.

\bibitem[{Cho et~al.(2021)Cho, Lei, Tan, and Bansal}]{cho2021vlt5}
Jaemin Cho, Jie Lei, Hao Tan, and Mohit Bansal. 2021.
\newblock Unifying vision-and-language tasks via text generation.
\newblock In \emph{ICML}.

\bibitem[{Dettmers et~al.(2023)Dettmers, Pagnoni, Holtzman, and
  Zettlemoyer}]{dettmers2023qlora}
Tim Dettmers, Artidoro Pagnoni, Ari Holtzman, and Luke Zettlemoyer. 2023.
\newblock {QLoRA}: Efficient finetuning of quantized {LLMs}.
\newblock \emph{arXiv preprint arXiv:2305.14314}.

\bibitem[{Devlin et~al.(2018)Devlin, Chang, Lee, and
  Toutanova}]{devlin2019bert}
Jacob Devlin, Ming-Wei Chang, Kenton Lee, and Kristina Toutanova. 2018.
\newblock {BERT}: Pre-training of deep bidirectional transformers for language
  understanding.
\newblock \emph{arXiv preprint arXiv:1810.04805}.

\bibitem[{Ferraro et~al.(2015)Ferraro, Mostafazadeh, Huang, Vanderwende,
  Devlin, Galley, and Mitchell}]{ferraro-etal-2015-survey}
Francis Ferraro, Nasrin Mostafazadeh, Ting-Hao Huang, Lucy Vanderwende, Jacob
  Devlin, Michel Galley, and Margaret Mitchell. 2015.
\newblock A survey of current datasets for vision and language research.
\newblock In \emph{EMNLP}.

\bibitem[{Google(2023)}]{reversegeocode2023}
Google. 2023.
\newblock \href
  {https://developers.google.com/maps/documentation/javascript/examples/geocoding-reverse}
  {Reverse geocoding}.
\newblock Accessed: 2023-06-22.

\bibitem[{{HuggingFace}(2023)}]{multiqahuggingface}
{HuggingFace}. 2023.
\newblock \href
  {https://huggingface.co/sentence-transformers/multi-qa-MiniLM-L6-cos-v1}
  {sentence-transformers/multi-qa-minilm-l6-cos-v1}.
\newblock Accessed: 2023-08-13.

\bibitem[{International(2018)}]{sae2018taxonomy}
Sae International. 2018.
\newblock Taxonomy and definitions for terms related to driving automation
  systems for on-road motor vehicles.
\newblock \emph{SAE international}.

\bibitem[{Kasneci et~al.(2023)Kasneci, Se{\ss}ler, K{\"u}chemann, Bannert,
  Dementieva, Fischer, Gasser, Groh, G{\"u}nnemann, H{\"u}llermeier
  et~al.}]{kasneci2023chatgpt}
Enkelejda Kasneci, Kathrin Se{\ss}ler, Stefan K{\"u}chemann, Maria Bannert,
  Daryna Dementieva, Frank Fischer, Urs Gasser, Georg Groh, Stephan
  G{\"u}nnemann, Eyke H{\"u}llermeier, et~al. 2023.
\newblock Chatgpt for good? on opportunities and challenges of large language
  models for education.
\newblock \emph{Learning and Individual Differences}.

\bibitem[{Large et~al.(2017)Large, Burnett, Antrobus, and
  Skrypchuk}]{driversfatigue}
David Large, Gary Burnett, Vicki Antrobus, and Lee Skrypchuk. 2017.
\newblock Stimulating conversation: Engaging drivers in natural language
  interactions with an autonomous digital driving assistant to counteract
  passive task-related fatigue.
\newblock In \emph{International Conference on Driver Distraction and
  Inattention}.

\bibitem[{Liang et~al.(2021)Liang, Wu, Morency, and
  Salakhutdinov}]{liang2021towards}
Paul~Pu Liang, Chiyu Wu, Louis-Philippe Morency, and Ruslan Salakhutdinov.
  2021.
\newblock Towards understanding and mitigating social biases in language
  models.
\newblock In \emph{ICML}.

\bibitem[{Lin(2004)}]{lin-2004-rouge}
Chin-Yew Lin. 2004.
\newblock {ROUGE}: A package for automatic evaluation of summaries.
\newblock In \emph{Text Summarization Branches Out}. ACL.

\bibitem[{Lin et~al.(2018)Lin, Hsu, Talamonti, Zhang, Oney, Mars, and
  Tang}]{10.1145/3242587.3242593}
Shih-Chieh Lin, Chang-Hong Hsu, Walter Talamonti, Yunqi Zhang, Steve Oney,
  Jason Mars, and Lingjia Tang. 2018.
\newblock Adasa: A conversational in-vehicle digital assistant for advanced
  driver assistance features.
\newblock In \emph{ACM Symposium on User Interface Software and Technology}.
  Association for Computing Machinery.

\bibitem[{Liu et~al.(2023)Liu, Han, Ma, Zhang, Yang, Tian, He, Li, He, Liu, Wu,
  Zhu, Li, Qiang, Shen, Liu, and Ge}]{liu2023summary}
Yiheng Liu, Tianle Han, Siyuan Ma, Jiayue Zhang, Yuanyuan Yang, Jiaming Tian,
  Hao He, Antong Li, Mengshen He, Zhengliang Liu, Zihao Wu, Dajiang Zhu, Xiang
  Li, Ning Qiang, Dingang Shen, Tianming Liu, and Bao Ge. 2023.
\newblock Summary of chatgpt/gpt-4 research and perspective towards the future
  of large language models.
\newblock \emph{arXiv preprint arXiv:2304.01852}.

\bibitem[{Lu et~al.(2021)Lu, Ding, Zhang, Li, Peng, and
  Liu}]{lu-etal-2021-engage}
Zexin Lu, Keyang Ding, Yuji Zhang, Jing Li, Baolin Peng, and Lemao Liu. 2021.
\newblock Engage the public: Poll question generation for social media posts.
\newblock In \emph{ACL-IJCNLP}.

\bibitem[{Mehta and Rastegari(2021)}]{mehta2022mobilevit}
Sachin Mehta and Mohammad Rastegari. 2021.
\newblock Mobilevit: light-weight, general-purpose, and mobile-friendly vision
  transformer.
\newblock \emph{arXiv preprint arXiv:2110.02178}.

\bibitem[{Mostafazadeh et~al.(2016)Mostafazadeh, Misra, Devlin, Mitchell, He,
  and Vanderwende}]{mostafazadeh-etal-2016-generating}
Nasrin Mostafazadeh, Ishan Misra, Jacob Devlin, Margaret Mitchell, Xiaodong He,
  and Lucy Vanderwende. 2016.
\newblock Generating natural questions about an image.
\newblock In \emph{ACL}.

\bibitem[{OpenAI(2023)}]{openai2023gpt4}
OpenAI. 2023.
\newblock {GPT-4 Technical Report}.
\newblock \emph{arXiv preprint arXiv:2303.08774}.

\bibitem[{Papineni et~al.(2002)Papineni, Roukos, Ward, and
  Zhu}]{papineni-etal-2002-bleu}
Kishore Papineni, Salim Roukos, Todd Ward, and Wei-Jing Zhu. 2002.
\newblock {B}leu: a method for automatic evaluation of machine translation.
\newblock In \emph{ACL}.

\bibitem[{Parekh et~al.(2022)Parekh, Poddar, Rajpurkar, Chahal, Kumar, Joshi,
  and Cho}]{electronics11142162}
Darsh Parekh, Nishi Poddar, Aakash Rajpurkar, Manisha Chahal, Neeraj Kumar,
  Gyanendra~Prasad Joshi, and Woong Cho. 2022.
\newblock A review on autonomous vehicles: Progress, methods and challenges.
\newblock \emph{Electronics}.

\bibitem[{Pratt et~al.(2020)Pratt, Yatskar, Weihs, Farhadi, and
  Kembhavi}]{Pratt2020Swig}
Sarah Pratt, Mark Yatskar, Luca Weihs, Ali Farhadi, and Aniruddha Kembhavi.
  2020.
\newblock Grounded situation recognition.
\newblock In \emph{ECCV}.

\bibitem[{Raffel et~al.(2020)Raffel, Shazeer, Roberts, Lee, Narang, Matena,
  Zhou, Li, and Liu}]{raffel2020exploring}
Colin Raffel, Noam Shazeer, Adam Roberts, Katherine Lee, Sharan Narang, Michael
  Matena, Yanqi Zhou, Wei Li, and Peter~J Liu. 2020.
\newblock Exploring the limits of transfer learning with a unified text-to-text
  transformer.
\newblock \emph{The Journal of Machine Learning Research}.

\bibitem[{Rajpurkar et~al.(2016)Rajpurkar, Zhang, Lopyrev, and
  Liang}]{rajpurkar2016squad}
Pranav Rajpurkar, Jian Zhang, Konstantin Lopyrev, and Percy Liang. 2016.
\newblock {SQ}u{AD}: 100,000+ questions for machine comprehension of text.
\newblock In \emph{EMNLP}.

\bibitem[{Ren et~al.(2015)Ren, He, Girshick, and Sun}]{fasterrcnn}
Shaoqing Ren, Kaiming He, Ross Girshick, and Jian Sun. 2015.
\newblock Faster {R-CNN:} towards real-time object detection with region
  proposal networks.
\newblock \emph{Advances in Neural Information Processing Systems}.

\bibitem[{Schwenk et~al.(2022)Schwenk, Khandelwal, Clark, Marino, and
  Mottaghi}]{schwenk2022aokvqa}
Dustin Schwenk, Apoorv Khandelwal, Christopher Clark, Kenneth Marino, and
  Roozbeh Mottaghi. 2022.
\newblock {A-OKVQA}: A benchmark for visual question answering using world
  knowledge.
\newblock \emph{arXiv preprint arXiv:2206.01718}.

\bibitem[{Sellam et~al.(2020)Sellam, Das, and
  Parikh}]{DBLP:journals/corr/abs-2004-04696}
Thibault Sellam, Dipanjan Das, and Ankur~P Parikh. 2020.
\newblock {BLEURT:} learning robust metrics for text generation.
\newblock \emph{arXiv preprint arXiv:2004.04696}.

\bibitem[{Sun et~al.(2020)Sun, Yu, Song, Liu, Yang, and
  Zhou}]{sun2020mobilebert}
Zhiqing Sun, Hongkun Yu, Xiaodan Song, Renjie Liu, Yiming Yang, and Denny Zhou.
  2020.
\newblock {MobileBERT}: a compact task-agnostic bert for resource-limited
  devices.
\newblock \emph{arXiv preprint arXiv:2004.02984}.

\bibitem[{Tay et~al.(2021)Tay, Dehghani, Rao, Fedus, Abnar, Chung, Narang,
  Yogatama, Vaswani, and Metzler}]{tay2022scale}
Yi~Tay, Mostafa Dehghani, Jinfeng Rao, William Fedus, Samira Abnar, Hyung~Won
  Chung, Sharan Narang, Dani Yogatama, Ashish Vaswani, and Donald Metzler.
  2021.
\newblock Scale efficiently: Insights from pre-training and fine-tuning
  transformers.
\newblock \emph{arXiv preprint arXiv:2109.10686}.

\bibitem[{Touvron et~al.(2023)Touvron, Lavril, Izacard, Martinet, Lachaux,
  Lacroix, Rozi{\`e}re, Goyal, Hambro, Azhar et~al.}]{touvron2023llama}
Hugo Touvron, Thibaut Lavril, Gautier Izacard, Xavier Martinet, Marie-Anne
  Lachaux, Timoth{\'e}e Lacroix, Baptiste Rozi{\`e}re, Naman Goyal, Eric
  Hambro, Faisal Azhar, et~al. 2023.
\newblock Llama: Open and efficient foundation language models.
\newblock \emph{arXiv preprint arXiv:2302.13971}.

\bibitem[{Vanderwende et~al.(2015)Vanderwende, Menezes, and
  Quirk}]{vanderwende-etal-2015-amr}
Lucy Vanderwende, Arul Menezes, and Chris Quirk. 2015.
\newblock An {AMR} parser for {E}nglish, {F}rench, {G}erman, {S}panish and
  {J}apanese and a new {AMR}-annotated corpus.
\newblock In \emph{NAACL}.

\bibitem[{Wang et~al.(2022)Wang, Yang, Men, Lin, Bai, Li, Ma, Zhou, Zhou, and
  Yang}]{wang2022ofa}
Peng Wang, An~Yang, Rui Men, Junyang Lin, Shuai Bai, Zhikang Li, Jianxin Ma,
  Chang Zhou, Jingren Zhou, and Hongxia Yang. 2022.
\newblock Ofa: Unifying architectures, tasks, and modalities through a simple
  sequence-to-sequence learning framework.
\newblock In \emph{ICML}.

\bibitem[{Wei et~al.(2022)Wei, Tay, Bommasani, Raffel, Zoph, Borgeaud,
  Yogatama, Bosma, Zhou, Metzler, Chi, Hashimoto, Vinyals, Liang, Dean, and
  Fedus}]{wei2022emergent}
Jason Wei, Yi~Tay, Rishi Bommasani, Colin Raffel, Barret Zoph, Sebastian
  Borgeaud, Dani Yogatama, Maarten Bosma, Denny Zhou, Donald Metzler, Ed~H.
  Chi, Tatsunori Hashimoto, Oriol Vinyals, Percy Liang, Jeff Dean, and William
  Fedus. 2022.
\newblock Emergent abilities of large language models.
\newblock \emph{TMLR}.

\bibitem[{Yan et~al.(2023)Yan, Sha, Zhao, Li, Martinez-Maldonado, Chen, Li,
  Jin, and Ga{\v{s}}evi{\'c}}]{yan2023practical}
Lixiang Yan, Lele Sha, Linxuan Zhao, Yuheng Li, Roberto Martinez-Maldonado,
  Guanliang Chen, Xinyu Li, Yueqiao Jin, and Dragan Ga{\v{s}}evi{\'c}. 2023.
\newblock Practical and ethical challenges of large language models in
  education: A systematic literature review.
\newblock \emph{arXiv preprint arXiv:2303.13379}.

\bibitem[{Yeh et~al.(2022)Yeh, Chen, Huang, and Ku}]{MVQG}
Min-Hsuan Yeh, Vincent Chen, Ting-Hao Huang, and Lun-Wei Ku. 2022.
\newblock Multi-{VQG}: Generating engaging questions for multiple images.
\newblock In \emph{EMNLP}.

\bibitem[{Zamir and Shah(2014)}]{6710175}
Amir~Roshan Zamir and Mubarak Shah. 2014.
\newblock Image geo-localization based on multiplenearest neighbor feature
  matching usinggeneralized graphs.
\newblock \emph{PAMI}.

\bibitem[{Zhang et~al.(2023)Zhang, Zhang, Pertsch, Liu, Ren, Chang, Sun, and
  Lim}]{zhang2023bootstrap}
Jesse Zhang, Jiahui Zhang, Karl Pertsch, Ziyi Liu, Xiang Ren, Minsuk Chang,
  Shao-Hua Sun, and Joseph~J Lim. 2023.
\newblock Bootstrap your own skills: Learning to solve new tasks with large
  language model guidance.
\newblock In \emph{Conference on Robot Learning}.

\bibitem[{Zhang and Choi(2021)}]{zhang2021situatedqa}
Michael~J.Q. Zhang and Eunsol Choi. 2021.
\newblock {S}ituated{QA}: Incorporating extra-linguistic contexts into {QA}.
\newblock \emph{EMNLP}.

\bibitem[{Zhang et~al.(2019)Zhang, Kishore, Wu, Weinberger, and
  Artzi}]{DBLP:journals/corr/abs-1904-09675}
Tianyi Zhang, Varsha Kishore, Felix Wu, Kilian~Q Weinberger, and Yoav Artzi.
  2019.
\newblock {BERTScore}: Evaluating text generation with {BERT}.
\newblock \emph{arXiv preprint arXiv:1904.09675}.

\end{thebibliography}
\bibliographystyle{acl_natbib}

\clearpage

\appendix





\section{Additional Diversity Analyses on GPT-4 Generated Questions}

We perform further diversity analyses on the questions provided in our LocaVQG dataset.

\begin{itemize}
    \item \textbf{Question type analysis}: While Table 3 shows the top 15 most frequent trigrams of generated questions, we have performed an additional trigram analysis during the rebuttal period to examine the diversity of the generated questions. In particular, we followed \citet{MVQG} and identified $2437$ question types among our $35$K generated questions. This highlights the diversity of the generated questions.
    \item \textbf{Pairwise cosine similarity}: Inspired by \citet{schwenk2022aokvqa}, which computes the average pairwise cosine similarity between each pair of generated questions encoded by a sentence transformer (multi-qa-MiniLM-L6-cos-v1 provided by~\citet{multiqahuggingface}) in a dataset, we have performed this evaluation on our generated dataset. We obtained an average cosine similarity of $0.1698$, indicating that the generated questions are not highly correlated and therefore ensuring the diversity of our proposed dataset.
\end{itemize}

\section{Latency Analysis}

Our goal is to develop lightweight models that can run on mobile devices. 
To examine the applicability of our proposed model FDT5 and the baselines, we measure and report the latency of MVQG-VL-T5, T5-Large, T5-Base, T5-Tiny, and our proposed FDT5 in \mytable{table:latency}. 
Each inference and post-filtering time is computed by averaging over $300$ trials to reduce the variance.

\settablecounter{15}
\begin{table}[h]
\centering
\resizebox{\linewidth}{!}{%
\begin{tabular}{l c c c }
    \toprule
    \textbf{Latency (sec)} & \textbf{Loading Model} & \textbf{Inference} & \textbf{Post-Filtering} \\
    \midrule
    MVQG-VL-T5 & $7.09$  & $6.38$ & N/A \\
    T5-Large & $12.79$  & $10.04$ & N/A \\
    T5-Base & $10.34$  & $5.9$ & N/A \\
    T5-Tiny & $3.89$  & $2.02$ & N/A \\
    FDT5 & $4.25$  & $2.27$ & $3.92$ \\
    \bottomrule
\end{tabular}}
\caption{\textbf{{Latency Testing of the trained models.}}}
\label{table:latency}
\end{table}

The results show that FDT5 and T5-Tiny, with the same model architecture and the same number of parameters, enjoy a significantly reduced time for loading models and running inference. The post-filtering phase of FDT5 takes $3.92$ seconds on average, indicating that the engaging question classifier requires FDT5 to perform $1.73$ additional inference trials for each LocaVQG task on average. Note that this post-filtering phase can be shut down for latency-critical scenarios, and FDT5 without post-filtering still outperforms T5-Tiny in human evaluation, according to ~\mytable{table:human_eval} and ~\mytable{table:ablation_post_filter}.

\section{Filtering Out Non-engaging Questions Generated by GPT-4 Using GPT-4}

This work proposes to train an engaging question classifier to filter out non-engaging questions generated by GPT-4;
alternatively, we can use GPT-4 to evaluate and filter out non-engaging questions that it generates.
To investigate this possibility, 
we feed the questions generated by GPT-4 back into GPT-4 for scoring (\ie determining if each generated question is engaging or not). 

Specifically, we provide GPT-4 with $10$ questions generated by itself and asked it to determine if each question is engaging or not. The prompt and the response of GPT-4 are shown in~\mytable{table:gpt_selfeval}. 
As a result, GPT-4 classifies all the questions it generated as engaging. This suggests that GPT-4 may not be useful to score its own generated questions, and motivates learning an external model to score the questions, such as our proposed engaging question classifier.

Note that this finding aligns with the statement, “we observe that GPT-4 assigns significantly higher scores to its own outputs compared to human ratings," stated in~\citet{dettmers2023qlora}.

\settablecounter{12}
\begin{table*}[h]
\centering
\scalebox{0.8}{%
\begin{tabular}{l}
    \toprule
    \textbf{Prompting GPT-4 for Self-Evaluation}\\
    \midrule
    You will be given 10 questions. Please classify this question into engaging and non-engaging questions. \\
    1. Did you know that Pittsburgh is known as the \"City of Bridges\" due to its 446 bridges? \\
    Can you spot any nearby bridges from this location? \\
    2. The tall buildings on the south side are a part of Pittsburgh's beautiful skyline. Can anyone name some of the prominent \\ buildings that make up the skyline? \\
    3. As we pass by these parked cars on the north side, can you guess the most popular car brand in Pittsburgh based on what \\ you see? \\
    4. Pittsburgh has a rich history of steel production, earning it the nickname \"Steel City.\" Can you see any indications of this \\ industrial past in the architecture of the buildings around us? \\
    5. The traffic light at the empty intersection on our east might make you wonder about the city's traffic patterns. What do \\you think contributes to the current flow of traffic in this area? \\
    6. Pittsburgh is known for its vibrant arts scene. Can anyone spot a nearby art gallery or theater as we drive along these city \\ streets? \\
    7. As we pass by the buildings on the west, let's talk about the city's diverse architectural styles. Can you identify different \\ styles just by looking at their exteriors? \\
    8. Pittsburgh is also famous for its rich sports culture, with successful teams in football, baseball, and hockey. Can you spot \\any sports fans or sports-themed establishments in our surroundings? \\
    9. The city has a growing tech scene, with companies like Google and Uber opening offices here. Can you see any tech \\ companies or startups as we drive through these streets? \\
    10. Pittsburgh is known for its friendly locals, often called \"yinzers\" due to their unique accent. Have you encountered any friendly \\ yinzers during your visit so far, and what has been your favorite interaction with them? \\
    \midrule
    \midrule
    \textbf{GPT-4 Response} \\
    \midrule
    1. Engaging \\
    2. Engaging \\
    3. Engaging \\
    4. Engaging \\
    5. Engaging \\
    6. Engaging \\
    7. Engaging \\
    8. Engaging \\
    9. Engaging \\
    10. Engaging \\
    \bottomrule
\end{tabular}}
\caption{\textbf{{GPT-4 Self Evaluation Results. }}}
\label{table:gpt_selfeval}
\end{table*}

\section{Effect of Incorporating Address Information}
\label{sec:app_address}
We are using the address as the main cue for GPT-4 to retrieve some information regarding those places.
While it is possible to produce general, non-location-specific questions based on hand-crafted templates, 
we found that GPT-4 can certainly produce location-specific questions that require knowledge of specific locations,
which can potentially be more engaging. Some examples are presented as follows.
\begin{itemize}
    \item Did you know that Fort Duquesne Boulevard is named after the historic Fort Duquesne, which was a key location during the French and Indian War? Have any of you studied that period in history?
    \item The city of Pittsburgh is known for its numerous bridges. How many bridges do you think are in the city, and why do you think there are so many?
    \item The 59th Street Bridge, also known as the Ed Koch Queensboro Bridge, connects Manhattan to Queens. Can you identify any famous movies or TV shows that have featured this iconic bridge?
\end{itemize}

We cannot obtain these questions by simply replacing the city name in other questions, nor should these questions be asked at a different location. 

Moreover, while generating a question with an address can increase the vocabulary size and average question length by simply inserting the address into the question, we still observe some questions that are generated based on the knowledge extracted by GPT-4 according to the address. We provide some example questions as follows.

\begin{itemize}
    \item \textbf{Address}: 1250 Penn Ave, Pittsburgh, PA 15222, USA 
        \begin{itemize}
            \item \textbf{Generated question}: “As we drive along 1250 Penn Ave, are there any upcoming events, festivals, or celebrations in the area that you'd like to learn more about?”
            \item \textbf{Observation}: GPT-4 knows this location has hosted several events in the past and therefore asks about upcoming events.
        \end{itemize}
    \item \textbf{Address}: 333 Boulevard of the Allies, Pittsburgh, PA 15222, USA 
        \begin{itemize}
            \item \textbf{Generated question}: Did you know that the Boulevard of the Allies is named to honor the Allies of World War I? What do you think about the significance of this historical connection?
            \item \textbf{Observation}: Based on knowing the history of the Boulevard of the Allies, GPT-4 asks about World War I.
        \end{itemize}
\end{itemize}

\section{Importance of Incorporating Visual Input and Learning to Generate Questions}
\label{sec:app_visual_learning}
Since it is possible to generate questions solely based on the fetched address, 
we aim to further analyze the effect of employing visual inputs to produce questions.
Also, as discussed in~\mysecref{sec:app_address},
we aim to quantitatively compare 
generating questions by our learned model and producing questions using general hand-crafted templates.
To this end, we labeled $100$ questions generated by FDT5 based on the following two criteria:
\begin{itemize}
    \item The generated questions contain visual information (w/ vis) or not (w/o vis)
    \item The generated questions are based on some templated (templated) or a learned language model (learned)
\end{itemize}
Specifically, we went through each question and
\begin{itemize}
    \item Determined if it contains visual information (e.g., describing surroundings). If so, this question is labeled as w/ vis; otherwise, it is labeled as w/o vis.
    \item Decided if it can be generated based on some templates (e.g., the city's name can be replaced with another city and the question still makes sense). If so, this question is labeled as templated; otherwise, it is labeled as learned.
\end{itemize}

Then, we analyze the engagement score and diversity of each group of questions. 
Regarding engagement scores, the questions containing visual information (w/ vis) achieve an average score of $4.007$, slightly outperforming the questions without visual information (w/o vis) with an average score of $3.915$, indicating that the visual-related questions may be more engaging. On the other hand, as the reviewer anticipated, the templated questions have a higher average score of $4.002$ compared to learned questions ($3.901$).

Then, we analyzed the diversity of each group of questions and found that the learned questions are more diverse (with a pairwise Cosine similarity score of $0.3614$), outperforming the templated questions with a pairwise Cosine similarity score of $0.3995$.

In conclusion, we believe that generating engaging and diverse questions still requires incorporating visual inputs with a well-learned language model.

\section{Implementation Details}
\label{sec:app_model}

\subsection{GPT-4 Parameters and Expenses}

\myparagraph{Setup}
When generating our proposed dataset using GPT-4, we use the model "\texttt{gpt4}" listed in the OpenAI API, with $0.7$ temperature and $0.1$ presence penalty.

\myparagraph{Expenses}
On average, each request of our task uses up around $500$ tokens, costing us around $\$ 0.001$ USD. 
In total, generating the dataset and experimenting with it
cost around $\$150 - 200$ USD.

\subsection{T5 and FDT5}

\myparagraph{Input} 
Similar to GPT-4, the T5 models and FDT5 take image captions as input.
The text input to the T5 models and FDT5 is modified from the chat prompt provided to GPT-4.
Specifically, we prepend "generate questions:" prefix to each input, resulting in the model input as: \textit{generate questions: You are currently driving on [Street Address]. On your North, [Image Caption]. On your East, [Image Caption]. On your South, [Image Caption]. On your West, [Image Caption]}.

\myparagraph{Implementation}
We adopt the basic pre-trained T5 models available on the Hugging Face platform.

\myparagraph{Training}
During the training, we use $5$ questions for each LocaVQG task tuple. We train each model for $20$ epochs with a learning rate of $10^{-4}$. 

\subsection{VL-T5}

\myparagraph{Input} 
The input of VL-T5 contains the following prompt prefix, street address, visual embeddings, and visual semantic groundings. 
\begin{itemize}
    \item \textbf{Prompt prefix.} The prompt prefix is  \textit{generate question:}, which guides the model to generate questions with the instruction. 
    
    \item \textbf{Street address.} The street address is the specific street address of the pictures that is verbalized, \eg \textit{You are currently driving in Penn Avenue, Pittsburgh}. 

    \item \textbf{Visual embeddings and visual semantic groundings.} We extract the visual embeddings from the whole image and the image regions with Faster-RCNN~\cite{fasterrcnn}. 
    Also, we adopted the grounded situation recognition (GSR)~\cite{Pratt2020Swig} model to extract information on the sequence of images to understand the semantics. Prefix of the directions of the images is also added, \eg \textit{North: [Visual embeddings]}
    
\end{itemize}

\myparagraph{Training}    
During the training, we used the pre-trained baselines presented in~\cite{MVQG}, specifically the adapter. 
We train the model for $30$ epochs with a learning rate of $10^{-5}$. We also employ gradient accumulation steps of $4$ and warm-up steps of $10$.

\subsection{Engaging Question Classifier}
The engaging question classifier is trained on the questions from the two datasets: SQuaD ($20239$ questions) and MVQG ($31098$ questions). The engaging question classifier is a BERT-based classifier with $110$M parameters. We train the classifier to classify the questions sampled from SQuaD as non-engaging and those sampled from MVQG as engaging for $10$ epochs. We use the ADAM optimizer with a learning rate of $10^{-5}$. 

Some example questions are as follows:
\begin{itemize}
    \item Why is that man playing billiards by himself? (Engaging)
    \item How did you celebrate your last birthday party? (Engaging)\
    \item What document was signed in 1999? (Non-engaging)
    \item What did John Milton do for world literature? (Non-engaging)
\end{itemize}

We report the performance of the learned engaging question classifier on the training (train), validation (val), and testing (test) sets in~\mytable{table:engaging_performance}. The accuracy evaluates if the classifier can correctly distinguish the questions in the MVQG dataset from those in the SQuaD dataset. 
The results show that the trained engaging question classifier can accurately distinguish the questions from the two datasets.

\settablecounter{14}
\begin{table} 
\centering
\scalebox{0.8}{%
\begin{tabular}{l c c c }
    \toprule
     & \textbf{train} & \textbf{valid} & \textbf{test} \\
    \midrule
    \textbf{Accuracy} & $99.9$\%  & $98.9$\% & $99.0$\% \\
    \bottomrule
\end{tabular}}
\caption{\textbf{{Performance of the Learned Engaging Question Classifier.}}}
\label{table:engaging_performance}
\end{table}


\section{Amazon Mechanical Turk Details}
\label{sec:app_amt}

\subsection{Human Evaluation Details}

\begin{figure*}[h]
    \centering
    \includegraphics[width=\linewidth]{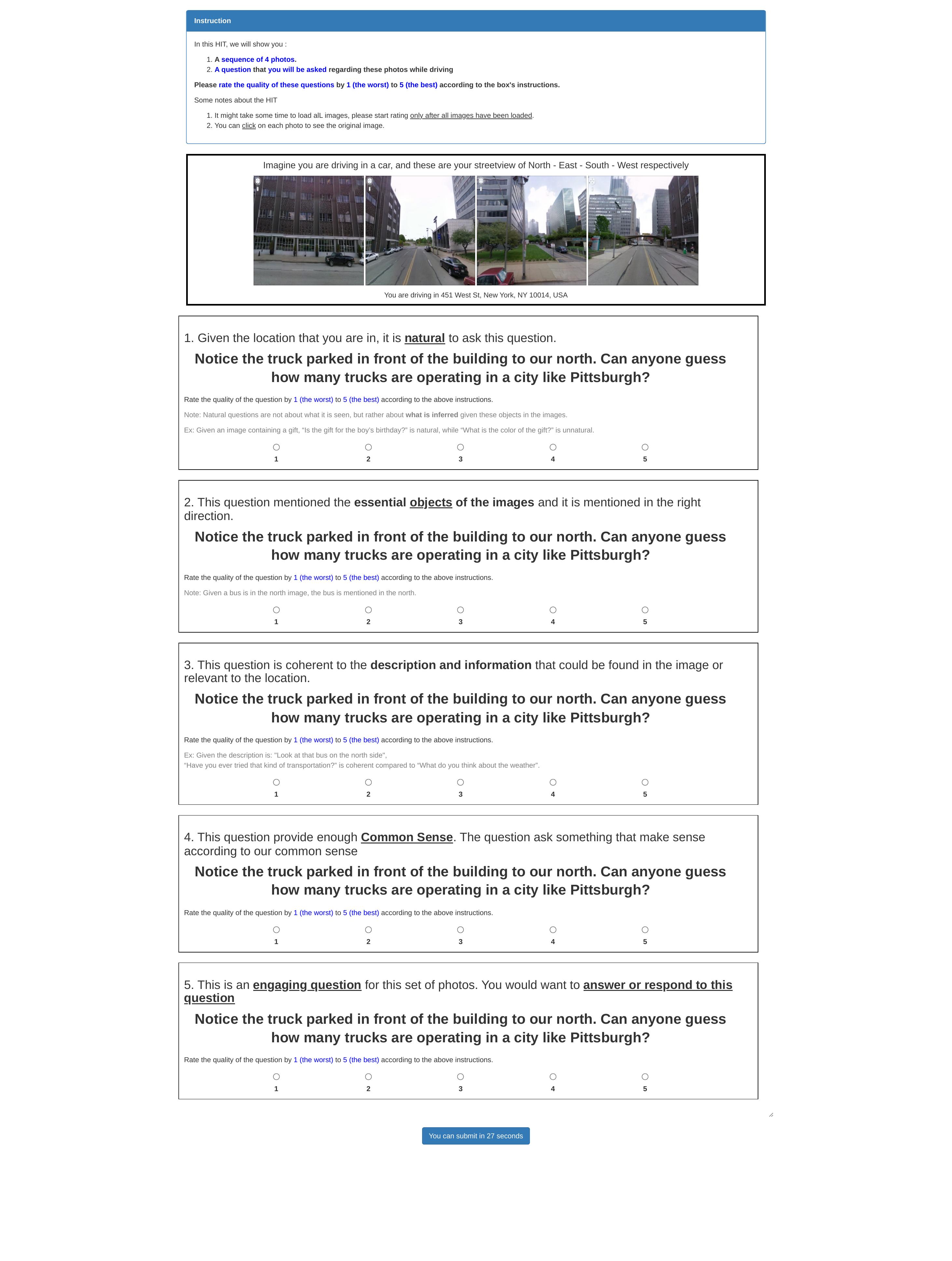}    
    \caption{
    \textbf{Human Evaluation AMT Interface.}
    }
    \label{fig:rating_survey}
\end{figure*}

\mysecref{sec:human_eval} conducts human evaluation on AMT to
compare questions generated by all models, GPT-4, and humans.
For each generated question, each worker is provided with the street address and $4$ streetview images of this location.
Then, the worker is required to rate the generated question according to $5$ evaluation metrics: Engagement, Naturalness, Coherence, Common Sense, and Grounding.
The descriptions of the metrics are as follows.
\begin{itemize}
    \item \textbf{Engagement} This is an engaging question for this set of photos. You would want to answer or respond to this question 
    \item \textbf{Naturalness} Given the pictures and location that you are in, it is natural to ask this question.
    \item \textbf{Coherence} This question asks something about the description and information that could be found in the image or relevant to the location.
    \item \textbf{Common Sense} This question provide enough Common Sense. The question asks something that makes sense according to our common sense.
    \item \textbf{Grounding} This question mentions the essential objects or information of the images, and it is mentioned in the right direction or talking about the location.
\end{itemize}
This AMT interface is shown in~\myfig{fig:rating_survey}

\subsection{Collecting Human Generated Questions}

\begin{figure*}[h]
    \centering
    \includegraphics[width=\linewidth]{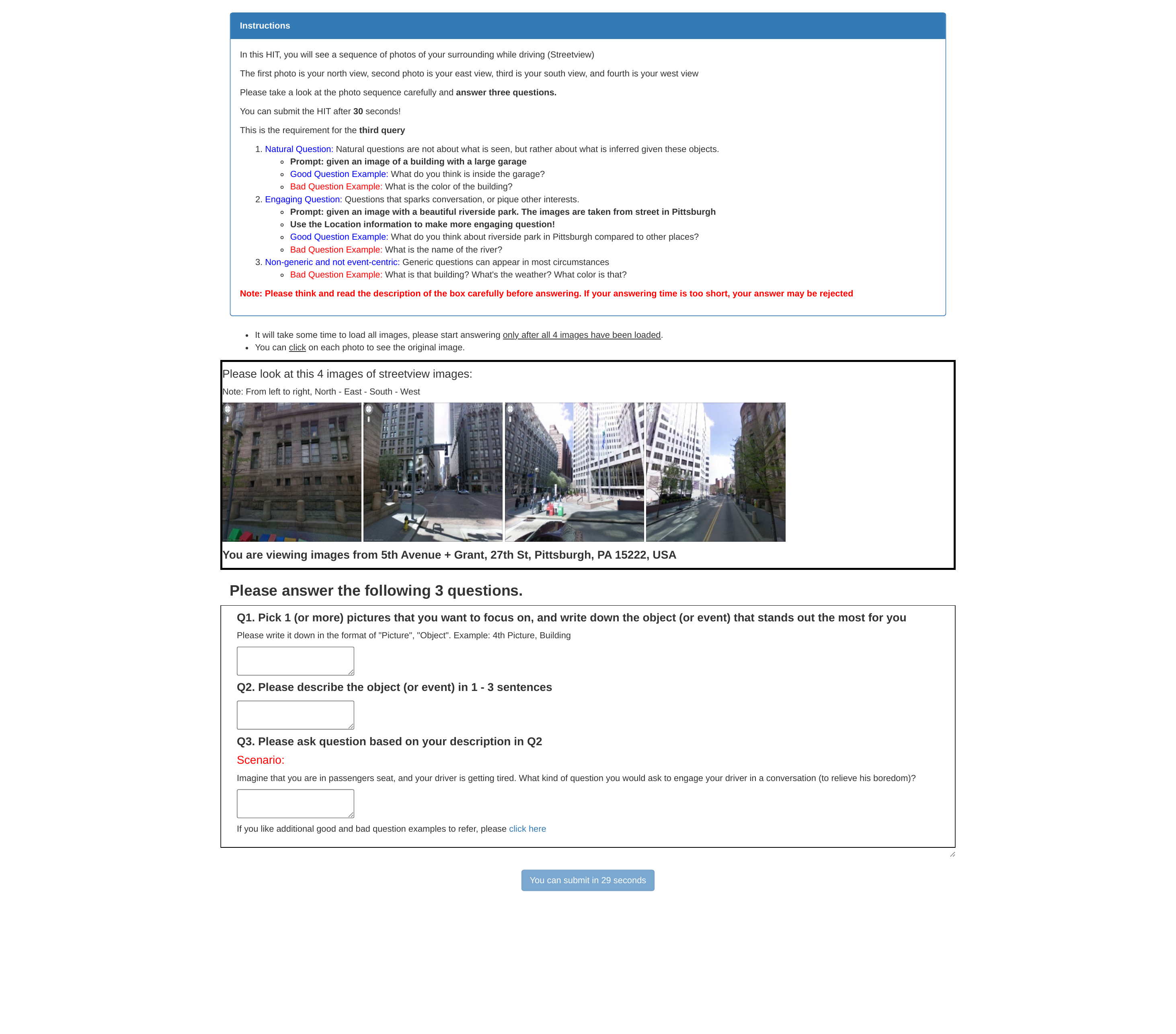}    
    \caption{
    \textbf{Location-aware Question Collection AMT Interface.}
    }
    \label{fig:annotation_survey}
\end{figure*}

In order to compare GPT-4 to humans regarding the ability to produce engaging questions from location-aware information,
we crowdsource human-generated questions from human annotators on AMT.
We use three-stage questions to collect questions as follows.
\begin{itemize}
    \item \textbf{Question 1.} Pick 1 (or more) pictures that you want to focus on, and write down the object (or event) that stands out the most for you.
    \item \textbf{Question 2.} Please describe the object (or event).
    \item \textbf{Question 3.} Please write a question based on it.
\end{itemize}
This AMT interface is shown in~\myfig{fig:annotation_survey}.
For each LocaVQG task tuple, we require three AMT workers to write a question, and then take the best question from the three questions. 
We present some questions produced by the workers in \mytable{table:human_questions}.

\settablecounter{14}

\begin{table*}[t]
\centering
\scalebox{0.85}{\begin{tabular}{l}
    \midrule
    There is a big strong building around us. Do you want to try come and visit it? \\
    \midrule
    Can you give me some tips to how to drive as like you? \\
    \midrule
    what do you think this building is used for? \\
    \midrule
    where are your coming from and which place you are going to visit? \\
    \midrule
    Look at the building around us, what do you think about the building? I find it looking really sturdy \\
    \midrule
    On the east side, you could see a cargo with few people around it. What do you think is inside the cargo? \\
    \midrule
    Have you ever wondered what is beyond that colossal vertical gateway adorned with glass windows of elegance? \\
    I wonder if it could it be an art gallery or perhaps a stock exchange? \\
    It is Pittsburgh ya know, the \"Steel City\" the possibilities are endless! \\
    \midrule
    What do you think about Garland Avenue compared to other streets in Orlando? \\
    \midrule
\end{tabular}}
\caption{\textbf{Human-Generated Questions.}
}
\label{table:human_questions}
\end{table*}


\end{document}